\documentclass[10pt,twocolumn,letterpaper]{article}

\usepackage{wacv}
\usepackage{times}
\usepackage{epsfig}
\usepackage{graphicx}
\usepackage{amsmath}
\usepackage{amssymb}
\usepackage{booktabs}
\usepackage{xcolor}
\usepackage{enumitem}
\usepackage{lipsum}
\usepackage[ruled,vlined,linesnumbered]{algorithm2e}
\usepackage{physics}
\usepackage{mathtools}
\usepackage{appendix}

\definecolor{codegreen}{rgb}{0.0,0.6,0.0}

\SetCommentSty{mycommfont}
\usepackage{float} 
\usepackage{nidanfloat}
\usepackage{caption}
\usepackage{datetime}

\makeatletter
\newcommand{\algorithmfootnote}[2][\footnotesize]{%
  \let\old@algocf@finish\@algocf@finish
  \def\@algocf@finish{\old@algocf@finish
    \leavevmode\rlap{\begin{minipage}{\linewidth}
    #1#2
    \end{minipage}}%
  }%
}
\makeatother

\usepackage[pagebackref=true,breaklinks=true,letterpaper=true,colorlinks,bookmarks=false]{hyperref}

\wacvfinalcopy 

\def\wacvPaperID{****} 

\title{BoT-SORT: Robust Associations Multi-Pedestrian Tracking}

\begin{document}

\author{
Nir Aharon\thanks{Corresponding author.} \hspace{1cm} Roy Orfaig \hspace{1cm} Ben-Zion Bobrovsky \vspace*{1mm}\\
School of Electrical Engineering, Tel-Aviv University \\
{\tt\small niraharon1@mai1.tau.ac.il \{royorfaig,bobrov\}@tauex.tau.ac.il}
}
\twocolumn[{%
\renewcommand\twocolumn[1][]{#1}%
\maketitle
\ifwacvfinal\thispagestyle{empty}\fi
\begin{center}
    \vspace{-5mm}
    \centering
    \captionsetup{type=figure}
    \includegraphics[width=1.02\textwidth]{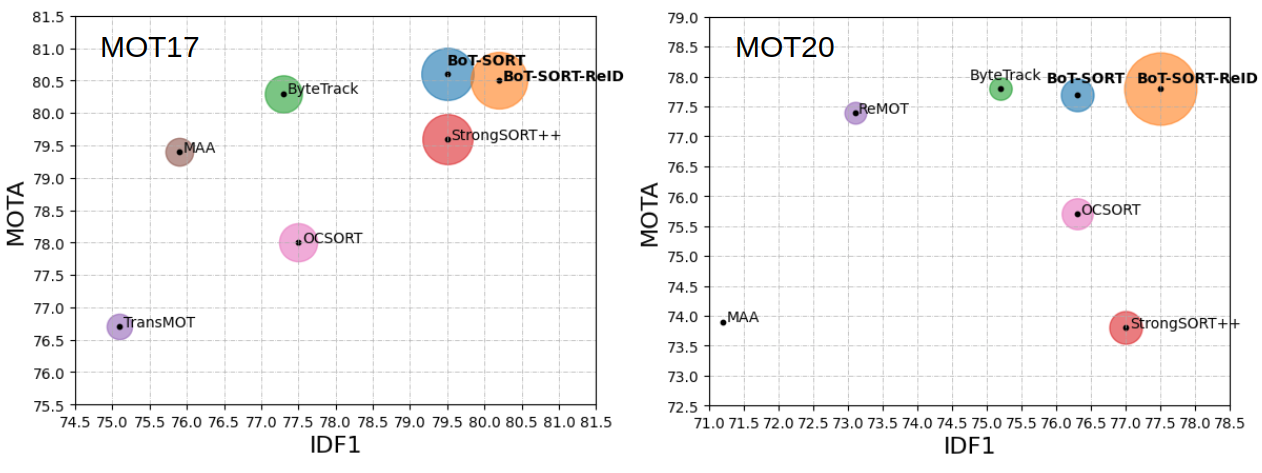}
    \captionof{figure}{IDF1-MOTA-HOTA comparisons of state-of-the-art trackers with our proposed BoT-SORT and BoT-SORT-ReID on the MOT17 and MOT20 test sets. The x-axis is IDF1, the y-axis is MOTA, and the radius of the circle is HOTA. Ours BoT-SORT-ReID and our simpler version BoT-SORT achieve the best IDF1, MOTA, and HOTA performance from all other trackers on the leadboards.}
    \label{fig:results_bubbles}
\end{center}%
}]

\begin{abstract}
The goal of multi-object tracking (MOT) is detecting and tracking all the objects in a scene, while keeping a unique identifier for each object. 
In this paper, we present a new robust state-of-the-art tracker, which can combine the advantages of motion and appearance information, along with camera-motion compensation, and a more accurate Kalman filter state vector. Our new trackers BoT-SORT, and BoT-SORT-ReID rank first in the datasets of MOTChallenge~\cite{milan2016mot16, dendorfer2020mot20} on both MOT17 and MOT20 test sets, in terms of all the main MOT metrics: MOTA, IDF1, and HOTA. For MOT17: 80.5 MOTA, 80.2 IDF1, and 65.0 HOTA are achieved. The source code and the pre-trained models are available at \url{https://github.com/NirAharon/BOT-SORT}

\end{abstract}

\section{Introduction}

Multi-object tracking (MOT) aims to detect and estimate the spatial-temporal trajectories of multiple objects in a video stream. MOT is a fundamental problem for numerous applications, such as autonomous driving, video surveillance, and more. 

Currently, tracking-by-detection has become the most effective paradigm for the MOT task ~\cite{yu2016poi, bewley2016simple, wojke2017simple, bochinski2017high, zhang2021bytetrack}. Tracking-by-detection contains an object detection step, followed by a tracking step.
The tracking step is usually built from two main parts: (1) Motion model and state estimation for predicting the bounding boxes of the tracklets in the following frames. A Kalman filter (KF)~\cite{brown1997introduction}, is the popular choice for this task.
(2) Associating the new frame detections with the current set of tracks. Two leading approaches are used for tackling the association task: (a) Localization of the object, mainly intersection-over-union (IoU) between the predicted tracklet bounding box and the detection bounding box. (b) Appearance model of the object and solving a re-identification (Re-ID) task. Both approaches are quantified into distances and used for solving the association task as a global assignment problem.

Many of the recent tracking-by-detection works based their study on the SORT~\cite{bewley2016simple},  DeepSORT~\cite{wojke2017simple} and JDE~\cite{wang2020towards} approaches. We have recognized some limitations in these "SORT-like" algorithms, which we will describe next.

Most SORT-like algorithms adopt the Kalman filter with the constant-velocity model assumption as the motion model. The KF is used for predicting the tracklet bounding box in the next frame for associating with the detection bounding box, and for predicting the tracklet state in case of occlusions or missed detections.

The use of the KF state estimation as the output for the tracker leads to a sub-optimal bounding box shape, compared to the detections driven by the object-detector. Most of the recent methods used the KF's state characterization proposed in the classic tracker DeepSORT~\cite{wojke2017simple}, which tries to estimate the aspect ratio of the box instead of the width, which leads to inaccurate width size estimations.

SORT-like IoU-based approaches mainly depend on the quality of the predicted bounding box of the tracklet. Hence, in many complex scenarios, predicting the correct location of the bounding box may fail due to camera motion, which leads to low overlap between the two related bounding boxes and finally to low tracker performance. We overcome this by adopting conventional image registration to estimate the camera motion, and properly correcting the Kalman filter. We denote this as Camera Motion Compensation (CMC).

Localization and appearance information (i.e. re-identification) within the SORT-like algorithms, in many cases lead to a trade-off between the tracker's ability to detect (MOTA) and the tracker's ability to maintain the correct identities over time (IDF1). Using IoU usually achieves better MOTA while Re-ID achieves higher IDF1.

In this work, we propose new trackers which outperform all leading trackers in all the main MOT metrics (Figure~\ref{fig:results_bubbles}) for the MOT17 and MOT20 challenges, by addressing the above SORT-like tracker's limitations and integrating them into the novel ByteTrack~\cite{zhang2021bytetrack}. In particular, the main contributions of our work can be summarized as follows:
\begin{itemize}[noitemsep,nolistsep]
\setlist[itemize]{align=parleft,left=0pt..1em}
\item We show that by adding improvements, such as a camera motion compensation-based features tracker and a suitable Kalman filter state vector for better box localization, tracking-by-detection trackers can be significantly improved.
\item We present a new simple yet effective method for IoU and ReID's cosine-distance fusion for more robust associations between detections and tracklets.
\end{itemize}

\section{Related Work} 
With the rapid improvements in object detection ~\cite{ren2015faster, duan2019centernet, redmon2018yolov3, bochkovskiy2020yolov4, ge2021yolox, zhu2020deformable} over the past few years, multi-object trackers have gained momentum. More powerful detectors lead to the higher tracking performance and reduce the need for complex trackers. Thus, tracking-by-detection trackers mainly focus on improving data association, while exploiting deep learning trends~\cite{zhang2021bytetrack, du2022strongsort}. \\

\noindent \textbf{Motion Models.}
Most of the recent tracking-by-detection algorithms are based on motion models. Recently, the famous Kalman filter~\cite{brown1997introduction} with constant-velocity model assumption, tends to be the popular choice for modeling the object motion~\cite{bewley2016simple, wojke2017simple, zhang2021fairmot, zhang2021bytetrack, han2022mat}. Many studies use more advanced variants of the KF, for example, the NSA-Kalman filter~\cite{du2021giaotracker, du2022strongsort}, which merges the detection score into the KF.
Many complex scenarios include camera motion, which may lead to non-linear motion of the objects and cause incorrect KF's predictions. Therefore, many researchers adopted camera motion compensation (CMC)~\cite{bergmann2019tracking, khurana2020detecting, han2022mat, stadler2022modelling, du2021giaotracker} by aligning frames via image registration using the Enhanced
Correlation Coefficient (ECC) maximization~\cite{evangelidis2008parametric} or matching features such as ORB~\cite{rublee2011orb}. \\

\noindent \textbf{Appearance models and re-identification.} Discriminating and re-identifying (ReID) objects by deep-appearance cues ~\cite{wang2018learning, zhou2019omni, Luo_2019_CVPR_Workshops} has also become popular, but falls short in many cases, especially when scenes are crowded, due to partial occlusions of persons. 
Separate appearance-based trackers crop the frame detection boxes and extract deep appearance features using an additional deep neural network~\cite{wojke2017simple, du2021giaotracker, du2022strongsort}. They enjoy advanced training techniques but demand high inference computational costs. 
Recently, several joint trackers \cite{wang2020towards, xu2020train, zhang2021fairmot, lu2020retinatrack, yu2021relationtrack, peng2020chained, liang2020rethinking, wang2021multiple}
have been proposed to train detection and some other components jointly, e.g., motion, embedding, and association models. The main benefit of these trackers is their low computational cost and comparable performance.

Lately, several recent studies ~\cite{stadler2022modelling, zhang2021bytetrack} have abandoned appearance information and relied only on high-performance detectors and motion information which achieve high running speed and state-of-the-art performance. In particular ByteTrack ~\cite{zhang2021bytetrack}, which exploits the low score detection boxes by matching the high confidence detections followed by another association with the low confident detections.\\ \\

\section{Proposed Method}
\begin{figure*}[t]
	\centering
	\hspace*{-7mm} 
	\includegraphics[scale=0.40]{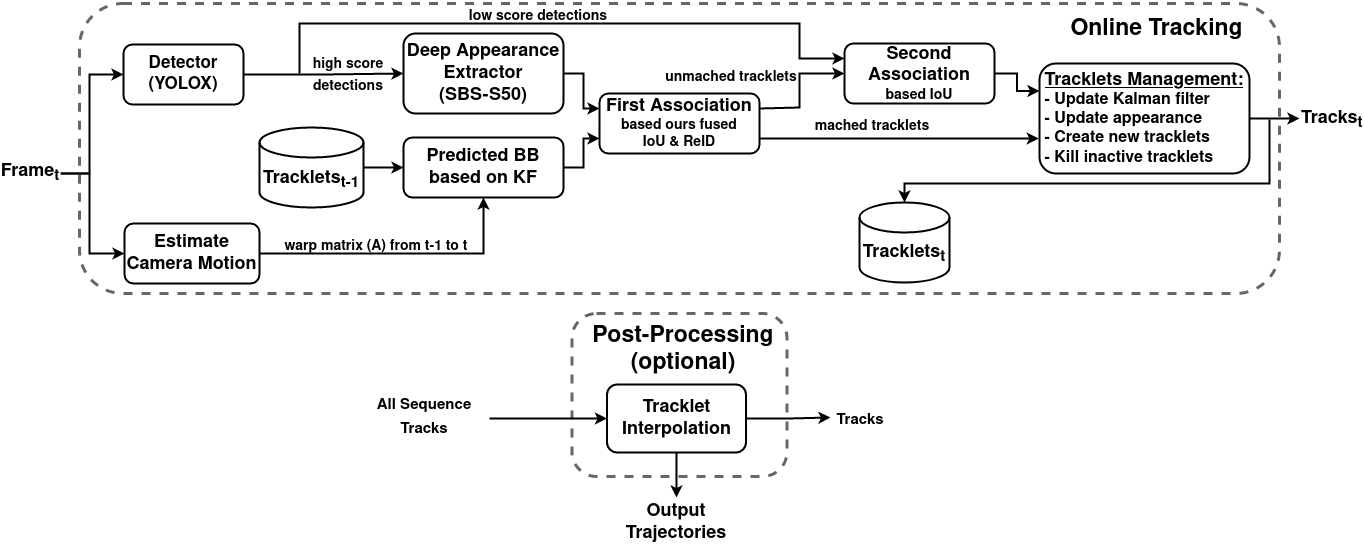}
	\caption{Overview of ours BoT-SORT-ReID tracker  pipeline. The online tracking region is the main part of ours tracker, and the post-processing region is optional addition, as in ~\cite{zhang2021bytetrack}.}
	\label{fig:alg_flow}
    \vspace{3mm}
\end{figure*}

In this section, we present our three main modifications and improvements for the multi-object tracking-based tracking-by-detection methods. By integrating these into the celebrated ByteTrack~\cite{zhang2021bytetrack}, we present two new state-of-the-art trackers, BoT-SORT and BoT-SORT-ReID.
BoT-SORT-ReID is a BoT-SORT extension including a re-identification module. Refer to Appendix~\ref{appendix:algorithm} for pseudo-code of ours BoT-SORT-ReID. The pipeline of our algorithm is presented in Fig~\ref{fig:alg_flow}.

\subsection{Kalman Filter}
To model the object's motion in the image plane, it is widely common to use the discrete Kalman filter with a constant-velocity model ~\cite{wojke2017simple}, see Appendix~\ref{appendix:kalman_filter} for details.

In SORT~\cite{bewley2016simple} the state vector was chosen to be a seven-tuple, $\vb*{x} =  [x_c, y_c, s, a, \dot{x_c}, \dot{y_c}, \dot{s}]^\top$, where ($x_c$, $y_c$) are the 2D coordinates of the object center in the image plane. $s$ is the bounding box scale (area) and $a$ is the bounding box aspect ratio. In more recent trackers ~\cite{wojke2017simple, wang2020towards, zhang2021fairmot, zhang2021bytetrack, du2022strongsort} the state vector has changed to an eight-tuple, $\vb*{x} =  [x_c, y_c, a, h, \dot{x_c}, \dot{y_c}, \dot{a}, \dot{h}]^\top$.
However, we found through experiments, that estimating the width and height of the bounding box directly, results in better performance. Hence, we choose to define the KF's state vector as in Eq.~\eqref{eqn:kf_x}, and the measurement vector as in Eq.~\eqref{eqn:kf_z}.
The matrices Q, R were chosen in SORT~\cite{bewley2016simple} to be time indepent, however in DeepSORT~\cite{wojke2017simple} it was suggested to choose Q, R as functions of some estimated elements and some measurement elements, as can be seen in their Github source code  \footnote{\url{https://github.com/nwojke/deep_sort}\label{footnote_deep_sort}}. Thus, using this choice of Q and R results in time-dependent $\vb*{Q}_k$ and $\vb*{R}_k$.
Following our changes in the KF's state vector, the process noise covariance $\vb*{Q}_k$ and measurement noise covariance $\vb*{R}_k$ matrices were modified, see Eq.~\eqref{eqn:kf_Q}, ~\eqref{eqn:kf_R}. \\Thus we have:
\begin{equation}
\label{eqn:kf_x}
\begin{split}
\vb*{x}_k = [x_{c}{\scriptstyle(k)}, y_{c}{\scriptstyle(k)}, w{\scriptstyle(k)}, h{\scriptstyle(k)}, \\        
\dot{x_{c}}{\scriptstyle(k)}, \dot{y_{c}}{\scriptstyle(k)}, \dot{w}{\scriptstyle(k)}, \dot{h}{\scriptstyle(k)}]^\top
\end{split}
\end{equation}
\begin{equation}
\label{eqn:kf_z}
\vb*{z}_k = [z_{x_c}{\scriptstyle(k)}, z_{y_c}{\scriptstyle(k)}, z_{w}{\scriptstyle(k)}, z_{h}{\scriptstyle(k)}]^\top
\end{equation}

\begin{equation}
\label{eqn:kf_Q}
\begin{split}
    \vb*{Q}_k = diag \big{(} (\sigma_{p} \hat{w}_{k-1|k-1})^2, (\sigma_{p} \hat{h}_{k-1|k-1})^2, \\
    (\sigma_{p} \hat{w}_{k-1|k-1})^2, (\sigma_{p} \hat{h}_{k-1|k-1})^2, \\
    (\sigma_{v} \hat{w}_{k-1|k-1})^2, (\sigma_{v} \hat{h}_{k-1|k-1})^2, \\ 
    (\sigma_{v} \hat{w}_{k-1|k-1})^2 ,(\sigma_{v} \hat{h}_{k-1|k-1})^2\big{)} 
\end{split}
\end{equation}

\begin{equation}
\label{eqn:kf_R}
\begin{split}
    \vb*{R}_k = diag\big{(}(\sigma_{m} \hat{w}_{k|k-1})^2, (\sigma_{m} \hat{h}_{k|k-1})^2, \\
    (\sigma_{m} \hat{w}_{k|k-1})^2, (\sigma_{m} \hat{h}_{k|k-1})^2\big{)} 
\end{split}
\end{equation}

\begin{figure}[hbt] 
	\centering
	\includegraphics[width=1\linewidth]{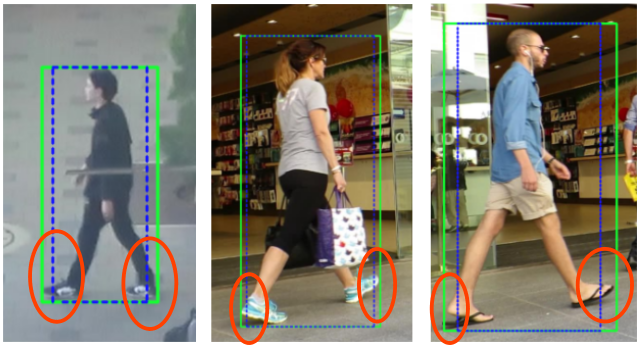}
	\caption{Visualization of the bounding box shape compare with the widely-used Kalman filter~\cite{wojke2017simple} (dashed blue) and the proposed Kalman filter (green). It seems that the bounding box width produced by the proposed KF fits more accurately to object. The dashed blue bounding box intersects the objects legs (in red), as the green bounding box reach to the desire width.}
	\label{fig:KF_width}
    \vspace{3mm}
\end{figure}

We choose the noise factors as in ~\cite{wojke2017simple} to be $\sigma_{p} = 0.05$, $\sigma_{v} = 0.00625$, and $\sigma_{m} = 0.05$, since our frame rate is also 30 FPS. Note, that we modified $\vb*{Q}$ and $\vb*{R}$ according to our slightly different state vector x.
In the case of track-loss, long predictions may result in box shape deformation, so proper logic is implemented, similar to ~\cite{zhang2021bytetrack}.
In the ablation study section, we show experimentally that those changes leads to higher HOTA. Strictly speaking, the reasons for the overall HOTA improvement is not clear to us. We assume that our modification of the KF contributes to improving the fit of the bounding box width to the object, as can be seen in Figure~\ref{fig:KF_width}.

\subsection{Camera Motion Compensation (CMC)}
\begin{figure*}[htp!]
	\centering
	\includegraphics[scale=0.66]{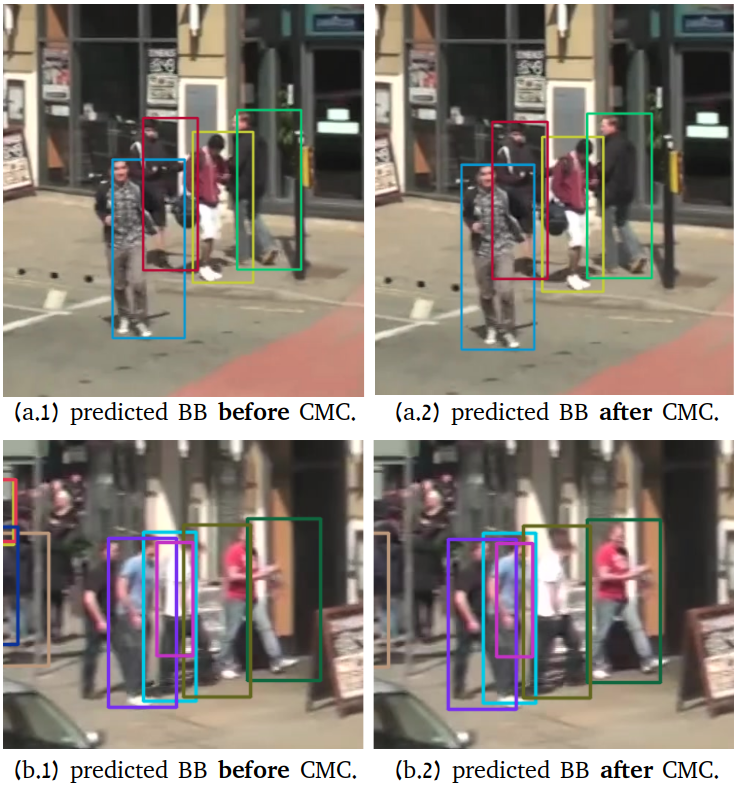}
	\caption{Visualization of the predicted tracklets bounding boxes, those predictions later used for association with the new detections BB based on maximum IoU criteria. 
	(a.1), (b.1) show the KF's predictions. (a.2), (b.2) show the KF's prediction after our camera motion compensation. 
	Figure (b.1) presents a scenario in which neglecting the camera motion will be expressed in IDSWs or FN. In contrast, in the opposite figure (b.2), the predictions fit their desirable location, and the association will succeed. The images are from the MOT17-13 sequence, which contains camera motion due to the vehicle turning right.} 
	\label{fig:KF_pred}
    \vspace{2mm}
\end{figure*}

Tracking-by-detection trackers rely heavily on the overlap between the predicted tracklets bounding boxes and the detected ones. In a dynamic camera situation, the bounding box location in the image plane can shift dramatically, which might result in increasing ID switches or false-negatives,  Figure~\ref{fig:KF_pred}. Trackers in static camera scenarios can also be affected due to motion by vibrations or drifts caused by the wind, as in MOT20, and in very crowded scenes ID-switches can be a real concern. The motion patterns in the video can be summarized as rigid motion, from the changing of the camera pose, and the non-rigid motion of the objects, e.g. pedestrians. With the lack of additional data on camera motion, (e.g. navigation, IMU, \textit{etc.}) or the camera intrinsic matrix, image registration between two adjacent frames is a good approximation to the projection of the rigid motion of the camera onto the image plane. We follow the global motion compensation (GMC) technique used in the OpenCV~\cite{opencv_library} implementation of the Video Stabilization module with affine transformation. This image registration method is suitable for revealing the background motion. First, extraction of image keypoints ~\cite{shi1994good} takes place, followed by sparse optical flow ~\cite{Bouguet1999PyramidalIO} for feature tracking with translation-based local outlier rejection. The affine matrix $\vb*{A}_{k-1}^k\in \mathbb{R}^{2\cross3}$ was solved using RANSAC ~\cite{fischler1981random}. The use of sparse registration techniques allows ignoring the dynamic objects in the scene based on the detections and thus having the potential of estimating the background motion more accurately.

For transforming the prediction bounding box from the coordinate system of frame $k-1$ to the coordinates of the next frame $k$, the calculated affine matrix $\vb*{A}_{k-1}^k$ was used, as will be described next.

The translation part of the transformation matrix only affects the center location of the bounding box, while the other part affects all the state vector and the noise matrix~\cite{white2019homography}. The camera motion correction step can be performed by the following equations: 

\begin{equation}
\begin{aligned}
    & \vb*{A}_{k-1}^k = 
    \begin{bmatrix}
        \vb*{M}_{2x2} | \vb*{T}_{2x1}
    \end{bmatrix} = 
    \begin{bmatrix}
        a_{11}\;a_{12}\;a_{13} \\ a_{21}\;a_{22}\;a_{23} \\ 
    \end{bmatrix}
\end{aligned}
\end{equation}
\begin{equation}
\begin{aligned}
{
    \vb*{\tilde{M}}}^k_{k-1} = 
    \begin{bmatrix}
    \vb*{M} & \vb*{0} & \vb*{0} & \vb*{0} \\
    \vb*{0} & \vb*{M} & \vb*{0} & \vb*{0} \\
    \vb*{0} & \vb*{0} & \vb*{M} & \vb*{0} \\
    \vb*{0} & \vb*{0} & \vb*{0} & \vb*{M}
    \end{bmatrix}, \; 
    \tilde{\vb*{T}}_{k-1}^k = 
    \begin{bmatrix}a_{13}\\a_{23}\\ 0\\ 0\\ \vdots \\ 0\end{bmatrix}
\end{aligned}
\end{equation}
\begin{equation}
\begin{aligned}
    \hat{\vb*{x}}^\prime_{k|k-1} = \tilde{\vb*{M}}_{k-1}^k \hat{\vb*{x}}_{k|k-1} + \tilde{\vb*{T}}_{k-1}^k \\
\end{aligned}
\end{equation}
\begin{equation}
    {{\vb*{P}}}^{\prime}_{k|k-1} = 
    \tilde{\vb*{M}}_{k-1}^k {\vb*{P}}_{k|k-1} {\tilde{\vb*{M}}_{k-1}}^{k^\top}
    \label{eq:cmc_cov}        
\end{equation}
When $\vb*{M}\in \mathbb{R}^{2\cross2}$ is a matrix containing the scale and rotations part of the affine matrix $A$, and $T$ contains the translation part. We use a mathematical trick by defining $\tilde{\vb*{M}}^k_{k-1}\in \mathbb{R}^{8\cross8}$ and $\tilde{\vb*{T}}_{k-1}^k\in \mathbb{R}^{8}$. Moreover, $\hat{\vb*{x}}_{k|k-1}$, $\hat{\vb*{x}}^\prime_{k|k-1}$ is the KF's predicted state vector at time $k$ before and after compensation of the camera motion respectively. ${\vb*P}_{k|k-1}$, ${{{\vb*{P}}}^{\prime}}_{k|k-1}$ is the KF's predicated covariance matrix before and after correction respectively. Afterwards, we use $\hat{\vb*{x}}^\prime_{k|k-1}$, ${{\hat{\vb*{P}}}^{\prime}}_{k|k-1}$ in the Kalman filter update step as follow:

\begin{equation}
    \begin{aligned}
    & \vb*{K_k} =  {{\vb*{P}}}^{\prime}_{k|k-1} \vb*{H}_k^\top (\vb*{H}_k  {{\vb*{P}}}^{\prime}_{k|k-1} \vb*{H}_k^\top + \vb*{R}_k)^{-1} \\
    & \hat{\vb*{x}}_{k|k} = \hat{\vb*{x}}^\prime_{k|k-1} + \vb*{K}_k (\vb*{z}_k - \vb*{H}_k \hat{\vb*{x}}^\prime_{k|k-1}) \\
    & \vb*{P}_{k|k} = (\vb*{I}- \vb*{K}_k \vb*{H}_k)  {{\vb*{P}}}^{\prime}_{k|k-1}
    \end{aligned}
    \label{eq:cmc_update_kf}
\end{equation} 
In high velocities scenarios, full correction of the state vector, including the velocities term, is essential. When the camera is changing slowly compared to the frame rate, the correction of Eq. \ref{eq:cmc_cov} can be omitted.
By applying this method our tracker becomes robust to camera motion. 

After compensating for the rigid camera motion, and under the assumption that the position of an object only slightly change from one frame to the next. In online high frame rate applications when missing detections occur, track extrapolations can be perform using the KF's prediction step, which may cause more continuous viewing of tracks with slightly higher MOTA. \\

\subsection{IoU - Re-ID Fusion}

To exploit the recent developments in deep visual representation, we integrated appearance features into our tracker. To extract these Re-ID features, we adopted the stronger baseline on top of BoT (SBS) \cite{Luo_2019_CVPR_Workshops} from the FastReID library, \cite{he2020fastreid} with the ResNeSt50 \cite{zhang2020resnest} as a backbone. We adopt the exponential moving average (EMA) mechanism for updating the matched tracklets appearance state $e_i^k$ for the $i$-th tracklet at frame $k$, as in~\cite{wang2020towards}, Eq.~\ref{eq:ema}.
\begin{equation}
    \begin{aligned}
    e_i^k = \alpha e_i^{k-1} + (1 - \alpha) f_i^k
    \end{aligned}
    \label{eq:ema}        
\end{equation}
Where $f_i^k$ is the appearance embedding of the current matched detection and $\alpha=0.9$ is a momentum term.
Because appearance features may be vulnerable to crowds, occluded and blurred objects, for maintaining correct feature vectors, we take into account only high confidence detections. For matching between the averaged tracklet appearance state $e_i^k$ and the new detection embedding vector $f_j^k$, cosine similarity is measured. We decided to abandon the common weighted sum between the appearance cost $A_a$ and motion cost $A_m$ for calculating the cost matrix $C$, Eq.~\ref{eq:weighted_sum}. 
\begin{equation}
C = \lambda A_a + (1 - \lambda) A_m, 
\label{eq:weighted_sum}
\end{equation}
Where the weight factor $\lambda$ is usually set to 0.98.

We developed a new method for combining the motion and the appearance information, \text{i.e.} the IoU distance matrix and the cosine distance matrix. First, low cosine similarity or far away candidates, in terms of IoU's score, are rejected. Then, we use the minimum in each element of the matrices as the final value of our cost matrix $C$. Our IoU-ReID fusion pipeline can be formulated as follows:
\begin{equation}
    \begin{aligned}
    \hat{d}_{i, j}^{cos} = 
        \begin{cases}
                0.5 \cdot d_{i, j}^{cos}, \text{($d_{i, j}^{cos} < \theta_{emb})$ $\land$ $(d_{i, j}^{iou} < \theta_{iou})$}\\
            1, \text{otherwise}
        \end{cases}
    \end{aligned}
    \label{eq:masked_dist}        
\end{equation}
\begin{equation}
    C_{i, j} = min\{d_{i, j}^{iou}, \hat{d}_{i, j}^{cos}\}
    \label{eq:min_dist}        
\end{equation}
Where $C_{i, j}$ is the $(i, j)$ element of cost matrix $C$. $d_{i, j}^{iou}$ is the IoU distance between tracklet $i$-th predicted bounding box and the $j$-th detection bounding box, representing the motion cost. $d_{i, j}^{cos}$ is the cosine distance between the average tracklet appearance descriptor $i$ and the new detection descriptor $j$. $\hat{d}_{i, j}^{cos}$ is our new appearance cost.
$\theta_{iou}$ is a proximity threshold, set to 0.5, used to reject unlikely pairs of tracklets and detections. $\theta_{emb}$ is the appearance threshold, which is used to separate positive association of tracklet appearance states and detections embedding vectors from the negatives ones.
We set $\theta_{emb}$ to 0.25 following Figure~\ref{fig:embedding_hist}.
The linear assignment problem of the high confidence detections \text{i.e.} first association step, was solved using the Hungarian algorithm~\cite{kuhn1955hungarian} and based on our cost matrix $C$, constructed with Eq.~\ref{eq:min_dist}.

\begin{figure}[htbp]
	\centering
    \hspace{-5mm}
	\includegraphics[width=1.0\linewidth]{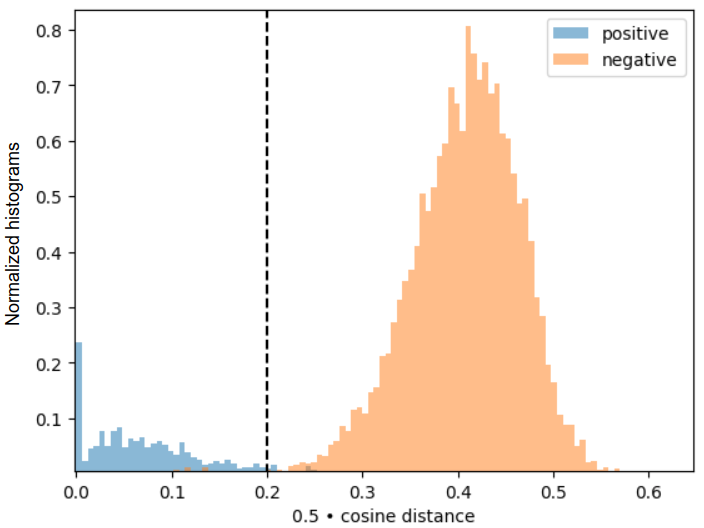}
	\caption{
	Study for the value of the appearance threshold $\theta_{emb}$ on the MOT17 validation set. FastReID's SBS-S50 model was trained on the first half of MOT17. 
	Positive indicated the same ID from different time and negative indicates different ID. It can be seen from the histogram that 0.25 is appropriate choice for $\theta_{emb}$. 
	} 
	\label{fig:embedding_hist}
    \vspace{2mm}
\end{figure}

\section{Experiments}
\subsection{Experimental Settings}
\noindent \textbf{Datasets.}
Experiments were conducted on two of the most popular benchmarks in the field of multi-object tracking for pedestrian detection and tracking in unconstrained environments: MOT17 \cite{milan2016mot16} and MOT20 \cite{dendorfer2020mot20} datasets under the ``private detection'' protocol. MOT17 contains video sequences filmed with both static and moving cameras. While MOT20 contains crowded scenes.
Both datasets contain training sets and test sets, without validation sets. For ablation studies, we follow \cite{zhou2020tracking, zhang2021bytetrack}, by using the first half of each video in the training set of MOT17 for training and the last half for validation.\\ 

\noindent \textbf{Metrics.} 
Evaluations were performed according to the widely accepted CLEAR metrics \cite{bernardin2008evaluating}. Including Multiple-Object Tracking Accuracy (MOTA), False Positive (FP), False Negative (FN), ID Switch (IDSW), \textit{etc.}, IDF1 \cite{ristani2016performance} and Higher-Order Tracking Accuracy (HOTA) \cite{luiten2021hota} to evaluate different aspects of the detection and tracking performance. Tracker speed (FPS) in Hz was also evaluated, although run time may vary significantly for different hardware.
MOTA is computed based on FP, FN, and IDSW. MOTA focuses more on detection performance because the amount of FP and FN are more significant than IDs. IDF1 evaluates the identity association performance. HOTA explicitly balances the effect of performing accurate detection, association, and localization into a single unified metric.\\

\noindent \textbf{Implementation details.}
All the experiments were implemented using PyTorch and ran on a desktop with 11th Gen Intel(R) Core(TM) i9-11900F @ 2.50GHz and NVIDIA GeForce RTX 3060 GPU.
For fair comparisons, we directly apply the publicly available detector of YOLOX \cite{ge2021yolox}, trained by ~\cite{zhang2021bytetrack} for MOT17, MOT20, and ablation study on MOT17. For the feature extractor, we trained FastReID's~\cite{he2020fastreid} SBS-50 model for MOT17 and MOT20 with their default training strategy, for 60 epochs. While we trained on the first half of each sequence and tested on the rest. The same tracker parameters were used throughout the experiments. The default detection score threshold $\tau$ was 0.6, unless otherwise specified. In the linear assignment step, if the detection and the tracklet similarity were smaller than 0.2, the matching was rejected. For the lost tracklets, we kept them for 30 frames in case it appeared again. Linear tracklet interpolation, with a max interval of 20, was performed to compensate for in-perfections in the ground truth, as in \cite{zhang2021bytetrack}.

\subsection{Ablation Study}
\noindent \textbf{Components Analysis.} Our ablation study mainly aims to verify the performance of our bag-of-tricks for MOT and quantify how much each component contributed. MOTChallenge official organization limits the number of attempts researchers can submit results to the test server. Thus, we used the MOT17 validation set, i.e. the second half of the train set of each sequence. To avoid the possible influence caused by the detector, we used ByteTrack's YOLOX-X MOT17 ablation study weights which were trained on CrowdHuman \cite{shao2018crowdhuman} and MOT17 first half of the train sequences. The same tracking parameters were used for all the experiments, as described in the Implementation details section. Table \ref{table_ablation} summarize the path from the outstanding ByteTrack to BoT-SORT and BoT-SORT-ReID. The Baseline represents our re-implemented ByteTrack, without any guidance from addition modules.\\
\begin{table*}[htbp]
\begin{center}
{
\begin{tabular}{cl | c c c c | c c c }
  \toprule
  & \textbf{Method} & \textbf{KF} & \textbf{CMC} & \textbf{Pred} & \textbf{w/ReID} 
  & \textbf{MOTA(↑)} & \textbf{IDF1(↑)} & \textbf{HOTA(↑)} \\
  \hline
  & Baseline (ByteTrack$^*$) & - & - & - & - & 77.66 & 79.77 & 67.88\\
  & Baseline + column 1 & \checkmark & & & & 77.67 & 79.89 & 68.12\\
  & Baseline + columns 1-2 & \checkmark & \checkmark & & & 78.31 & 81.51 & 69.06 \\
  & Baseline + columns 1-3 (BoT-SORT) & \checkmark & \checkmark & \checkmark & & 78.39 & 81.53 & 69.11 \\
  & Baseline + columns 1-4 (BoT-SORT-ReID) & \checkmark & \checkmark & \checkmark & \checkmark &\textbf{78.46} & \textbf{82.07} & \textbf{69.17} \\
  \bottomrule
\end{tabular}
}
\end{center}
\vspace{-2mm}
\raggedright \footnotesize{\textbf{$^*$}Ours reproduced results using TrackEval~\cite{luiten2020trackeval} with tracking threshold of 0.6, new track threshold of 0.7 and, first association matching threshold of 0.8.}
\caption
{
    Ablation study on the MOT17 validation set for basic strategies, i.e., Updated Kalman filter (KF), camera motion compensation (CMC), output tracks prediction (Pred), with ReID module (w/ReID). All results obtained with the same parameters set. (best in bold). 
}
\label{table_ablation}
\end{table*}

\noindent \textbf{Re-ID module.} 
Appearance descriptors are an intuitive way of associating the same person over time. It has the potential to overcome the large displacement and long occlusions. Most recent attempts using Re-ID with cosine similarity, outperform simply using IoU for high frame rate videos case. In this section, we compare different strategies for combining the motion and the visual embedding in the first matching association step of our tracker, on the MOT17 validation set, Table~\ref{table_reid_ablation}. IoU alone outperforms the Re-ID-based methods, excluding our proposed method. Hence, for low resources applications, IoU is a good design choice. Ours IoU-ReID combination with IoU masking achieves the highest results in terms of MOTA, IDF1, and HOTA, and benefits from the motion and the appearance information.\\

\begin{table*}[htbp!]
\begin{center}
{\setlength{\tabcolsep}{10.0pt}

\begin{tabular}{cl | c c c | c c c }
  \toprule
  & \textbf{Similarity} & \textbf{IoU} & \textbf{w/Re-ID} & \textbf{Masking}  & \textbf{MOTA(↑)} & \textbf{IDF1(↑)} & \textbf{HOTA(↑)} \\
  \hline
  & IoU & \checkmark &  &  & 78.4 & 81.5 & 69.1\\
  & Cosine & & \checkmark & & 73.7 & 70.0 & 62.4\\
  & JDE~\cite{wang2020towards} &   & \checkmark & Motion(KF) & 77.7 & 80.1 & 68.2 \\
  & Cosine &  & \checkmark & IoU & 78.3 & 81.0 & 68.7 \\
  & \textbf{Ours} & \checkmark & \checkmark & IoU &\textbf{78.5} & \textbf{82.1} & \textbf{69.2} \\
  \bottomrule
\end{tabular}

}
\end{center}
\vspace{-2mm}
\caption
{
    Ablation study on the MOT17 validation set for different similarities strategies for exploit the ReID module (w/ReID). Masking indicates strategy for rejecting distant associations. Ours proposed minimum between the IoU and the cosine achieve the highest scores (best in bold). 
}
\label{table_reid_ablation}
\end{table*}

\noindent \textbf{Online vs Offline.} 
Many applications are required to analyze events retrospectively. In these cases, the use of offline methods, such as global-link~\cite{du2022strongsort}, can significantly improve the results. In this study, we only focus on improving the online part of the tracker. For fair comparisons in the MOTChallenge benchmarks, we use simple linear tracklet interpolation, as in~\cite{zhang2021bytetrack}.\\ 

\noindent \textbf{Current MOTA.} 
One of the challenges of developing a multi-object tracker is to identify tracker failures using the standard MOT metrics. In many cases, finding the specific reasons or even the time range for the tracker failure can be time-consuming.
Hence, for analyzing the fall-backs and difficulties of multi-object trackers we evolve the MOTA metric to time or frame-dependent MOTA, which we call Current-MOTA (cMOTA). $\textit{cMOTA}(t_k)$ is simply the MOTA from $t=t_0$ to $t=t_k$. e.g. $\textit{cMOTA}(T)$ is equal to the classic MOTA, calculated over all the sequences, where $T$ is the sequence length, Eq.~\ref{eq:cmota}. 
\begin{equation}
    \textit{cMOTA}(t = T) = \textit{MOTA}
    \label{eq:cmota}        
\end{equation}

\begin{figure*}[htbp!]
	\centering
	\includegraphics[width=0.95\linewidth]{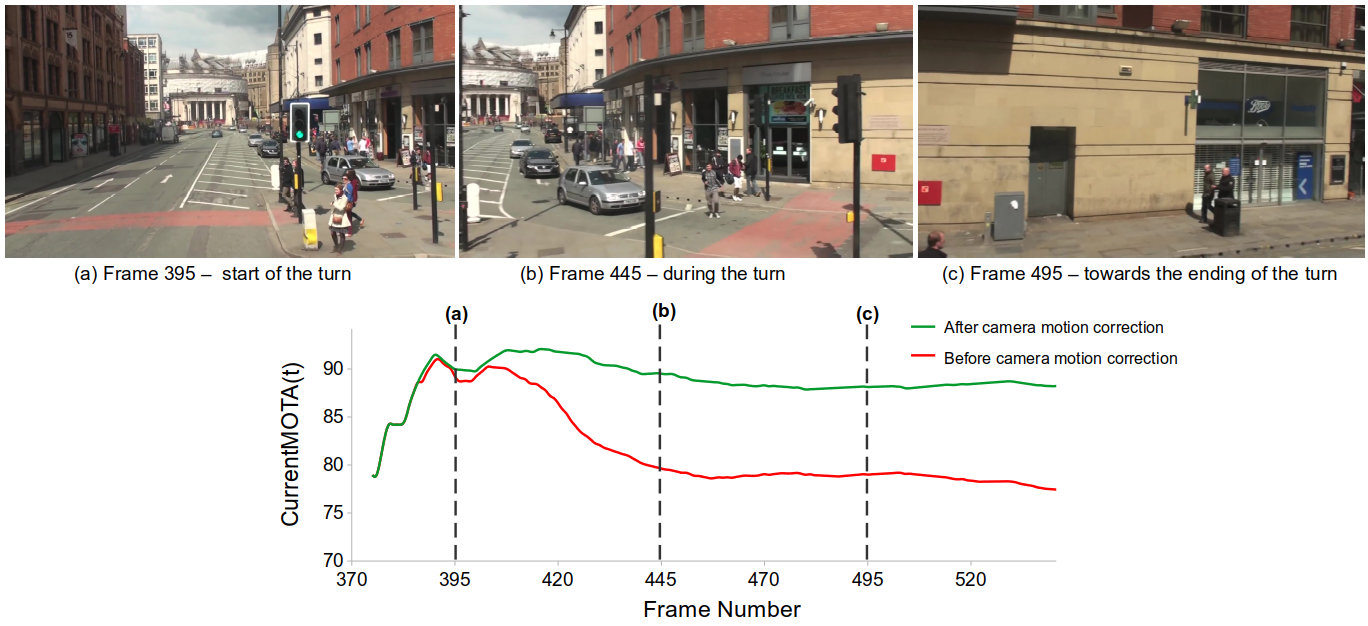}
	\caption{Example of the advantage of current-MOTA (cMOTA) graph in rotating camera scene from MOT17-13 validation set. By examine the cMOTA graph, one can see (in red), that the MOTA drops rapidly from frame 400 to 470 and afterward the cMOTA reach plateau. In this case, by looking at the suspected frames which the cMOTA reveals, we can detect that the reason for the tracker failure is the rotation of the camera. By adding ours CMC (in green), the MOTA  preserved high using the same detections and tracking parameters.}
	\label{fig:cMOTA}
    \vspace{1mm}
\end{figure*}

This allows us to easily find cases where the tracker fails. The same procedure can be replay with any of the CLEAR matrices, e.g. IDF1, \textit{etc.}. An example of the advantage of cMOTA can be found in Figure \ref{fig:cMOTA}. Potentially, cMOTA can help to identify and explore many other tracker failure scenarios. 

\subsection{Benchmarks Evaluation}
We compare our BoT-SORT and BoT-SORT-ReID state-of-the-art trackers on the test set of MOT17 and MOT20 under the private detection protocol in Table~\ref{table_mot17}, Table~\ref{table_mot20}, respectively. All the results are directly obtained from the official MOTChallenge evaluation server. Comparing FPS is difficult because the speed claimed by each method depends on the devices they are implemented on, and the time spent on detections is generally excluded for tracking-by-detection trackers.\\ \\

\noindent \textbf{MOT17.} BoT-SORT-ReID and the simpler version, BoT-SORT, both outperform all other state-of-the-art trackers in all main metrics, i.e. MOTA, IDF1, and HOTA. BoT-SORT-ReID is the first tracker to achieve IDF1 above 80, Table~\ref{table_mot17}. The high IDF1 along with the high MOTA in diverse scenarios indicates that our tracker is robust and effective. \\ \\
\begin{table*}[hbtp!]
\begin{center}
\scalebox{0.95}{\setlength{\tabcolsep}{10.0pt}

\begin{tabular}{ l | c | c | c | c | c | c | c} 

\toprule
Tracker & MOTA$\uparrow$ & IDF1$\uparrow$ & HOTA$\uparrow$ & FP$\downarrow$ & FN$\downarrow$ & IDs$\downarrow$ & FPS$\uparrow$ \\
\midrule

Tube\_TK \cite{pang2020tubetk}              & 63.0 & 58.6 & 48.0 & 27060 & 177483 & 4137 & 3.0\\
MOTR \cite{zeng2021motr}                    & 65.1 & 66.4 & - & 45486 & 149307 & 2049 & -\\
CTracker \cite{peng2020chained}             & 66.6 & 57.4 & 49.0 & 22284 & 160491 & 5529 & 6.8\\
CenterTrack \cite{zhou2020tracking}         & 67.8 & 64.7 & 52.2 & 18498 & 160332 & 3039 & 17.5\\
QuasiDense \cite{pang2021quasi}             & 68.7 & 66.3 & 53.9 & 26589 & 146643 & 3378 & 20.3\\
TraDes \cite{wu2021track}                   & 69.1 & 63.9 & 52.7 & 20892 & 150060 & 3555 & 17.5\\
MAT \cite{han2022mat}                       & 69.5 & 63.1 & 53.8 & 30660 & 138741 & 2844 & 9.0\\
SOTMOT \cite{zheng2021improving}            & 71.0 & 71.9 & - & 39537 & 118983 & 5184 & 16.0\\
TransCenter \cite{xu2021transcenter}        & 73.2 & 62.2 & 54.5 & 23112 & 123738 & 4614 & 1.0\\
GSDT \cite{wang2020joint}                   & 73.2 & 66.5 & 55.2 & 26397 & 120666 & 3891 & 4.9\\
Semi-TCL \cite{li2021semi}                  & 73.3 & 73.2 & 59.8 & 22944 & 124980 & 2790 & -\\
FairMOT \cite{zhang2021fairmot}             & 73.7 & 72.3 & 59.3 & 27507 & 117477 & 3303 & 25.9\\
RelationTrack \cite{yu2021relationtrack}    & 73.8 & 74.7 & 61.0 & 27999 & 118623 & 1374 & 8.5\\
PermaTrackPr \cite{tokmakov2021learning}    & 73.8 & 68.9 & 55.5 & 28998 & 115104 & 3699 & 11.9\\
CSTrack \cite{liang2020rethinking}          & 74.9 & 72.6 & 59.3 & 23847 & 114303 & 3567 & 15.8\\
TransTrack \cite{sun2020transtrack}         & 75.2 & 63.5 & 54.1 & 50157 & 86442 & 3603 & 10.0\\
FUFET \cite{shan2020tracklets}              & 76.2 & 68.0 & 57.9 & 32796 & 98475 & 3237 & 6.8\\
SiamMOT \cite{liang2021one}                 & 76.3 & 72.3 & - & - & - & - & 12.8\\
CorrTracker \cite{wang2021multiple}         & 76.5 & 73.6 & 60.7 & 29808 & 99510 & 3369 & 15.6\\
TransMOT \cite{chu2021transmot}             & 76.7 & 75.1 & 61.7 & 36231 & 93150 & 2346 & 9.6\\
ReMOT \cite{yang2021remot}                  & 77.0 & 72.0 & 59.7 & 33204 & 93612 & 2853 & 1.8\\
MAATrack \cite{stadler2022modelling}        & 79.4 & 75.9 & 62.0 & 37320 & 77661 & 1452 & \textbf{189.1}\\
OCSORT \cite{cao2022observation}            & 78.0 & 77.5 & 63.2 & \textbf{15129} & 107055 & 1950 & 29.0\\
StrongSORT++ \cite{du2022strongsort}        & 79.6 & 79.5 & 64.4 & 27876 & 86205 & \textbf{1194} & 7.1\\ 
ByteTrack \cite{zhang2021bytetrack}         & 80.3 & 77.3 & 63.1 & 25491 & \textbf{83721} & 2196 & 29.6 \\ \hline
\textbf{BoT-SORT (ours)}                    & \textbf{80.6} & 79.5 & 64.6 & 22524 & 85398 & 1257 & 6.6 \\
\textbf{BoT-SORT-ReID (ours)}               & 80.5 & \textbf{80.2} & \textbf{65.0} & 22521 & 86037 & 1212 & 4.5 \\
\bottomrule
\end{tabular}

}
\end{center}
\vspace{-2mm}
\caption{Comparison of the state-of-the-art methods under the “private detector” protocol on MOT17 test set. The best results are shown in \textbf{bold}. BoT-SORT and BoT-SORT-ReID ranks 2st and 1st respectively among all the MOT20 leadboard trackers.}
\label{table_mot17}
\end{table*}

\noindent \textbf{MOT20.}
MOT20 is considered to be a difficult benchmark due to crowded scenarios and many occlusion cases. Even so, BoT-SORT-ReID ranks 1st in terms of MOTA, IDF1 and HOTA, Table~\ref{table_mot20}. Some other trackers were able to achieve the same results in one metric (e.g. the same MOTA or IDF1) but their other results were compromised. Our methods were able to significantly improve the IDF1 and the HOTA while preserving the MOTA.\\
\begin{table*}[htb]
\begin{center}
\scalebox{0.95}{\setlength{\tabcolsep}{10.0pt}

\begin{tabular}{ l | c | c | c | c | c | c | c} 

\toprule
Tracker & MOTA$\uparrow$ & IDF1$\uparrow$ & HOTA$\uparrow$ & FP$\downarrow$ & FN$\downarrow$ & IDs$\downarrow$ & FPS$\uparrow$\\
\midrule
MLT \cite{zhang2020multiplex}               & 48.9 & 54.6 & 43.2 & 45660 & 216803 & 2187 & 3.7\\
FairMOT \cite{zhang2021fairmot}             & 61.8 & 67.3 & 54.6 & 103440 & 88901 & 5243 & 13.2\\
TransCenter \cite{xu2021transcenter}        & 61.9 & 50.4 & - & 45895 & 146347 & 4653 & 1.0\\
TransTrack \cite{sun2020transtrack}         & 65.0 & 59.4 & 48.5 & 27197 & 150197 & 3608 & 7.2\\
CorrTracker \cite{wang2021multiple}         & 65.2 & 69.1 & - & 79429 & 95855 & 5183 & 8.5\\
Semi-TCL \cite{li2021semi}                  & 65.2 & 70.1 & 55.3 & 61209 & 114709 & 4139 & -\\
CSTrack \cite{liang2020rethinking}          & 66.6 & 68.6 & 54.0 & 25404 & 144358 & 3196 & 4.5\\
GSDT \cite{wang2020joint}                   & 67.1 & 67.5 & 53.6 & 31913 & 135409 & 3131 & 0.9\\
SiamMOT \cite{liang2021one}                 & 67.1 & 69.1 & - & - & - & - & 4.3\\
RelationTrack \cite{yu2021relationtrack}    & 67.2 & 70.5 & 56.5 & 61134 & 104597 & 4243 & 2.7\\
SOTMOT \cite{zheng2021improving}            & 68.6 & 71.4 & - & 57064 & 101154 & 4209 & 8.5\\
MAATrack \cite{stadler2022modelling}        & 73.9 & 71.2 & 57.3 & 24942 & 108744 & 1331 & 14.7\\
OCSORT \cite{cao2022observation}            & 75.7 & 76.3 & 62.4 & 19067 & 105894 & 942 & \textbf{18.7}\\
StrongSORT++ \cite{du2022strongsort}        & 73.8 & 77.0 & 62.6 & \textbf{16632} & 117920 & \textbf{770} & 1.4\\ 
ByteTrack \cite{zhang2021bytetrack}         & \textbf{77.8} & 75.2 & 61.3 & 26249 & 87594 & 1223 & 17.5\\ \hline
\textbf{BoT-SORT (ours)}                    & 77.7 & 76.3 & 62.6 & 22521 & \textbf{86037} & 1212 & 6.6 \\
\textbf{BoT-SORT-ReID (ours)}               & \textbf{77.8} & \textbf{77.5} & \textbf{63.3} & 24638 & 88863 & 1257 & 2.4 \\ 

\bottomrule
\end{tabular}
}
\end{center}
\vspace{-2mm}
\caption{Comparison of the state-of-the-art methods under the “private detector” protocol on MOT20 test set. The best results are shown in \textbf{bold}. BoT-SORT and BoT-SORT-ReID ranks 2st and 1st respectively among all the MOT17 leadboard trackers.}
\label{table_mot20}
\end{table*}

\subsection{Limitations}
BoT-SORT and BoT-SORT-ReID still have several limitations.
In scenes with a high density of dynamic objects, the estimation of the camera motion may fail due to lack of background keypoints. Wrong camera motion may lead to unexpected tracker behavior.
Another real-life application concern is the run time.
Calculating the global motion of the camera can be time-consuming when large images need to be processed. But GMC run time is negligible compared to the detector inference time. Thus, multi-threading can be applied to calculating GMC, without any additional delays.
Separated appearance trackers have relatively low running speed compared with joint trackers and several appearance-free trackers. We apply deep feature extraction only for high confidence detections to reduce the computational cost.
If necessary, the feature extractor network can be merged into the detector head, in a joint-detection-embedding manner.

\section{Conclusion}
In this paper, we propose an enhanced multi-object tracker with MOT bag-of-tricks for a robust association, named BoT-SORT, which ranks 1st in terms of MOTA, IDF1, and HOTA on MOT17 and MOT20 datasets among all other trackers on the leadboards. This method and its components can easily be integrated into other tracking-by-detection trackers. In addition, a new MOT investigation tool - cMOTA is introduced. We hope that this work will help to push forward the multiple-object tracking field.

\section{Acknowledgement}
We thank Shlomo Shmeltzer Institute for Smart Transportation in Tel-Aviv University for their generous support of our Autonomous Mobile Laboratory.

{\small
\bibliographystyle{ieee}
\bibliography{egbib}
}

\newpage
\appendixtitleon
\appendixtitletocon
\begin{appendices}

\section{Pseudo-code of BoT-SORT-ReID}
\label{appendix:algorithm}

\begin{algorithm}[h!]
\DontPrintSemicolon
\SetNoFillComment
\SetCustomAlgoRuledWidth{0.95\linewidth}
\footnotesize
\KwIn{A video sequence $\texttt{V}$; object detector $\texttt{Det}$; appearance (features) extractor $\texttt{Enc}$; high detection score threshold {$\tau$}; new track score threshold {$\eta$}}
\KwOut{Tracks $\mathcal{T}$ of the video}

Initialization: $\mathcal{T} \leftarrow \emptyset$\;
\For{frame $f_k$ in $\texttt{V}$}{
	
	\tcc{Handle new detections}
	$\mathcal{D}_k \leftarrow \texttt{Det}(f_k)$\; 
	$\mathcal{D}_{high} \leftarrow \emptyset$\;
	$\mathcal{D}_{low} \leftarrow \emptyset$\;
	$\mathcal{F}_{high} \leftarrow \emptyset$\;
    \BlankLine	
	
	\For{$d$ in $\mathcal{D}_k$}{
	\If{$d.score > \tau$}{
	\tcc{Store high scores detections}
	$\mathcal{D}_{high} \leftarrow  \mathcal{D}_{high} \cup \{d\}$ \;
    \BlankLine	
	\tcc{Extract appearance features}
	$\mathcal{F}_{high} \leftarrow  \mathcal{F}_{high} \cup \texttt{Enc}(f_k, d.box)$ \;
	}
	\Else{
	\tcc{Store low scores detections}
	$\mathcal{D}_{low} \leftarrow  \mathcal{D}_{low} \cup \{d\}$ \;
	}
	}
    \BlankLine	
	\tcc{Find warp matrix from k-1 to k}
	$\mathcal{A}_{k-1}^k = findMotion(f_{k-1}, f_k)$ \;
	
    \BlankLine	
	\BlankLine
	\tcc{Predict new locations of tracks}
	\For{$t$ in $\mathcal{T}$}{
	$t \leftarrow \texttt{KalmanFilter}(t)$ \;
	$t \leftarrow \texttt{MotionCompensation}(t, \mathcal{A}_{k-1}^k)$ \;
	}
    
    \BlankLine
	\tcc{First association}
	
    $C_{iou} \leftarrow IOUDist(\mathcal{T}.boxes, \mathcal{D}_{high})$ \;
    $C_{emb} \leftarrow FusionDist(\mathcal{T}.features, \mathcal{F}_{high}, C_{iou})$ \text{\textcolor{codegreen}{// Eq.~\ref{eq:masked_dist}}} \; 
    
	$C_{high} \leftarrow min(C_{iou}, C_{emb})\;$\text{\textcolor{codegreen}{// Eq.~\ref{eq:min_dist}}}
	\;
    \BlankLine
    $ \text{Linear assignment by  Hungarian's alg. with}\;C_{high}$ \;
	\BlankLine
	$\mathcal{D}_{remain} \leftarrow \text{remaining object boxes from } \mathcal{D}_{high}$ \;
	$\mathcal{T}_{remain} \leftarrow \text{remaining tracks from } \mathcal{T}$ \;
	
	\BlankLine
    \BlankLine
    \tcc{Second association}
    $C_{low} \leftarrow IOUDist(\mathcal{T}_{remain}.boxes, \mathcal{D}_{low})$ \;
    $ \text{Linear assignment by  Hungarian's alg. with}\;C_{low}$ \;
	$\mathcal{T}_{re-remain} \leftarrow \text{remaining tracks from } \mathcal{T}_{remain}$ \;
	
	\BlankLine
	\BlankLine
	\tcc{Update matched tracks}
	\text{Update matched tracklets Kalman filter.} \;
	\text{Update tracklets appearance features.} \;
	
    \BlankLine
	\BlankLine
	\tcc{Delete unmatched tracks}
	$\mathcal{T} \leftarrow \mathcal{T} \setminus \mathcal{T}_{re-remain}$ \;
	
    \BlankLine
	\BlankLine
	\tcc{Initialize new tracks}
    \For{$d$ in $\mathcal{D}_{remain}$}{
    \If{$d.score > \eta$}{
	$\mathcal{T} \leftarrow  \mathcal{T} \cup \{d\}$ \;
	}
	}
}
\BlankLine
\tcc{(Optional) Offline post-processing}
$\mathcal{T} \leftarrow  LinearInterpolation(\mathcal{T})$ \;
\BlankLine
Return: $\mathcal{T}$
\caption{Pseudo-code of BoT-SORT-ReID.}
\algorithmfootnote{Remark: tracks rebirth~\cite{zhang2021bytetrack} is not shown in the algorithm for simplicity.}
\label{alg:bot_sort_reid}
\end{algorithm}

\section{Kalman Filter Model}
\label{appendix:kalman_filter}

The Kalman filter ~\cite{brown1997introduction} goal is to try to estimate the state $\vb*{x}\in \mathbb{R}^{n}$ given the measurements $\vb*{z}\in \mathbb{R}^{m}$ and given a known $\vb*{x}_0$, as $k \in \mathbb{N}^+$. In the task of object tracking, where no active control exists, the discrete-time Kalman filter is governed by the following linear stochastic difference equations:
 
\begin{equation}
\begin{aligned}
    & \vb*{x}_{k} = \vb*{F}_k \vb*{x}_{k-1}
     + \vb*{n}_{k-1}
\end{aligned}
\label{eq:kf_difference_state}
\end{equation}
\begin{equation}
\begin{aligned}
    & \vb*{z}_{k} = \vb*{H}_k \vb*{x}_{k}
     + \vb*{v}_{k}
\end{aligned}
\label{eq:kf_difference_noise}
\end{equation}

Where $\vb*{F}_k$ is the transition matrix from discrete-time $k-1$ to $k$. The observation matrix is $\vb*{H}_k$.
The random variables $\vb*{n}_{k}$ and $\vb*{v}_{k}$ represent the process and measurement noise respectively.
They are assumed to be independent and identically
distributed (i.i.d) with normal distribution.

\begin{equation}
\begin{aligned}
    & \vb*{n}_{k} \sim \mathcal{N}(\vb*{0}, \vb*{Q}_k), & \vb*{v}_{k} \sim \mathcal{N}(\vb*{0}, \vb*{R}_k)
\end{aligned}
\label{eq:noise_distributions}
\end{equation}

The process noise covariance $\vb*{Q}_k$ and measurement noise covariance $\vb*{R}_k$ matrices might change with each time step.
Kalman filter consists of a prediction step and update step. The entire Kalman filter can be summarized in the following recursive equations:
\begin{equation}
\begin{aligned}
    & \hat{\vb*{x}}_{k|k-1} = \vb*{F}_k \hat{\vb*{x}}_{k-1|k-1}\\
    & \vb*{P}_{k|k-1} = \vb*{F}_k \vb*{P}_{k-1|k-1} \vb*{F}_k^\top + \vb*{Q}_k
\end{aligned}
\label{eq:predict_kf}
\end{equation}

\begin{equation}
    \begin{aligned}
    & \vb*{K_k} = \vb*{P}_{k|k-1} \vb*{H}_k^\top (\vb*{H}_k \vb*{P}_{k|k-1} \vb*{H}_k^\top + \vb*{R}_k)^{-1} \\
    & \hat{\vb*{x}}_{k|k} = \hat{\vb*{x}}_{k|k-1} + \vb*{K}_k (\vb*{z}_k - \vb*{H}_k \hat{\vb*{x}}_{k|k-1}) \\
    & \vb*{P}_{k|k} = (\vb*{I}- \vb*{K}_k \vb*{H}_k) \vb*{P}_{k|k-1}
    \end{aligned}
    \label{eq:update_kf}
\end{equation} 

For proper choice of initial condition $\hat{\vb*{x}}_{0}$ and $\vb*{P}_{0}$, see literature, e.g.~\cite{brown1997introduction}, and more specifically refer to ~\cite{zhang2021bytetrack}.

At each step $k$, KF predicts the prior estimate of state $\hat{\vb*{x}}_{k|k-1}$ and the covariance matrix $\vb*{P}_{k|k-1}$. KF updates the posterior state estimation $\hat{\vb*{x}}_{k|k}$ given the observation $\vb*{z}_k$  and the estimated covariance $\vb*{P}_{k|k}$, calculated based on the optimal Kalman gain $\vb*{K}_k$. 

The constant-velocity model matrices corresponding to the state vector and measurement vector defined in Eq.~\ref{eqn:kf_x} and Eq.~\ref{eqn:kf_z}, present in Eq.~\ref{eq:const_vel_model}.
\begin{equation}
\label{eq:const_vel_model}
\begin{aligned}
    & \vb*{F} = 
    \begin{bmatrix}
        1\;0\;0\;0\;1\;0\;0\;0\\
        0\;1\;0\;0\;0\;1\;0\;0\\
        0\;0\;1\;0\;0\;0\;1\;0\\
        0\;0\;0\;1\;0\;0\;0\;1\\
        0\;0\;0\;0\;1\;0\;0\;0\\
        0\;0\;0\;0\;0\;1\;0\;0\\
        0\;0\;0\;0\;0\;0\;1\;0\\
        0\;0\;0\;0\;0\;0\;0\;1\\
    \end{bmatrix}, 
    & \vb*{H} = 
    \begin{bmatrix}
        1\;0\;0\;0\;0\;0\;0\;0\\
        0\;1\;0\;0\;0\;0\;0\;0\\
        0\;0\;1\;0\;0\;0\;0\;0\\
        0\;0\;0\;1\;0\;0\;0\;0\\
    \end{bmatrix}
\end{aligned}
\end{equation}

\end{appendices}

\end{document}